\documentclass[11pt]{article}

\usepackage[final]{acl}

\usepackage{times}
\usepackage{latexsym}

\usepackage[T1]{fontenc}

\usepackage[utf8]{inputenc}

\usepackage{microtype}

\usepackage{inconsolata}

\usepackage{graphicx}

 \usepackage{amsmath}
\usepackage{hyperref}
\usepackage{url}
\usepackage{algorithm}
\usepackage{algorithmic}
\usepackage[table]{xcolor} 
\usepackage{subcaption}
\usepackage{booktabs}
\usepackage{subcaption}
\usepackage{multirow}

\title{SyRHM: Symbolic-Language-Enhanced Reasoning with \\ Associative Retrieval for Zero-shot Harmful Meme Detection}

\author{
  \textbf{%
  Hanling Wang\textsuperscript{1}\thanks{Equal contribution.},
  Chenlong Wei\textsuperscript{2}\footnotemark[1],
  Yingjuan Li\textsuperscript{3},
  Di Wi\textsuperscript{2},
  }\\
  \textbf{%
  Yuchao Zhang\textsuperscript{4},
  Xiaohui Zhu\textsuperscript{2},
  Yao Zhu\textsuperscript{5}\thanks{Corresponding author.}
  }
\\
\textsuperscript{1} University of California, San Diego
\textsuperscript{2} Xi'an Jiaotong-Liverpool University
\\
\textsuperscript{3} University of Washington
\textsuperscript{4} The Australian National University
\textsuperscript{5} Zhejiang University
\\
\texttt{haw168@ucsd.edu}
\quad
\texttt{xiaohui.zhu@xjtlu.edu.cn}
\quad
\texttt{ee\_zhuy@zju.edu.cn}
}

\begin{document}
\maketitle
\begin{abstract}
Detecting harmful memes is critical for maintaining safe online communities. However, harmful intent is often implicit, arising from visual–textual incongruity and cultural stereotypes, which challenges existing multimodal detectors.
We propose SyRHM, a framework that decomposes harmful meme detection into meaning-grounded retrieval and symbolic-language-enhanced multi-stage reasoning. 
SyRHM retrieves semantically related memes by parsing multimodal content into textual elements and descriptions, providing grounded context beyond surface-level similarity. 
Building on the retrieved context, SyRHM uses a translator stage to convert multimodal inputs into symbolic intermediate representations, and then performs multi-stage reasoning via planner and solver stages, enabling expressive and interpretable analysis of harmful intent.
Experiments on FHM, HarM, and MultiOff demonstrate the effectiveness of SyRHM, achieving superior performance on most evaluation settings against multimodal and reasoning-based baselines, while providing reasoning traces for harmful content.

The code is available at: \url{https://github.com/Scabbards1500/SyRHM}

\textcolor{red}{\textbf{Disclaimer:} This paper contains offensive content that may be disturbing to some readers.}
\end{abstract}

\section{Introduction}
Memes have emerged as a powerful form of online expression, combining visual elements with text to convey complex, often culturally embedded messages. Beyond their humorous or entertaining purposes, memes can also serve as vehicles for covertly spreading harmful ideologies, posing a serious threat to inclusive online communities~\citep{sharma2022detecting, pramanick2021detecting, gomez2020exploring}. 
Detecting harmful memes~\citep{lippe2020multimodal, kiela2020hateful, cao2020deephate} is particularly challenging because harmfulness often lies not in explicit surface features but in implicit semantic associations, cultural references, or sarcasm. 
Therefore, effective detection requires both multimodal understanding and a reasoning process capable of inferring hostility from subtle, indirect cues.

\begin{figure}[!h]
\centering
\includegraphics[width=0.48\textwidth]{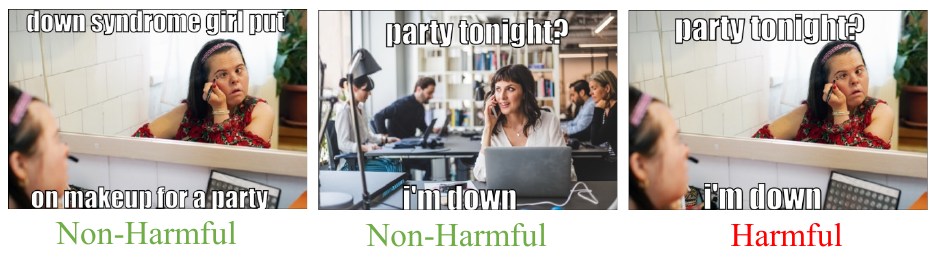} 
\caption{Similar memes can convey different meanings: the left examples are neutral, while the right example conveys harmful stereotypes targeting individuals with Down syndrome.} 
\label{img:compare}
\end{figure}

Early work on harmful meme detection primarily relied on supervised multimodal models~\citep{pramanick2021detecting, kiela2020hateful, suryawanshi2020multimodal} or fine-tuned vision–language models~\citep{lippe2020multimodal, muennighoff2020vilio, velioglu2020detecting, hee2022explaining}, which require large-scale high-quality annotations and struggle to generalize to evolving meme semantics.
To reduce annotation dependence, recent studies explore few-shot in-context learning for harmful meme detection~\citep{hee2024bridging, huang2024towards, cao2024modularized}. More recently, large multimodal models (LMMs) have been applied in zero-shot settings, often augmented with retrieval of related memes to provide contextual grounding~\citep{mei2024improving, qu2023evolution}.
Some approaches incorporate reasoning strategies, such as Chain-of-Evolution prompting in Evolver~\citep{huang2025evolver} and multi-agent debate in MinD~\citep{liu2025mind}, further improving decision robustness.

Despite these advances, retrieval-augmented methods suffer from two core limitations.
First, global image–text retrieval favors surface similarity over semantic relevance, 
often returning visually or textually similar memes with opposite meanings (Figure~\ref{img:compare}).
Second, reasoning over retrieved evidence relies only on surface-level cues from images and captions, without capturing semantic relations, thereby limiting the understanding of implicit harmful content.
To address these challenges, we propose SyRHM, a framework that integrates fine-grained associative retrieval with multi-stage symbolic-language-enhanced reasoning for zero-shot harmful meme detection. 
Instead of relying on holistic image--caption embeddings, the associative retrieval module in SyRHM parses each meme into structured semantic elements (e.g., entities, attributes, and relations) and textual descriptions, and retrieves semantically related meme neighbors at the element level rather than via global similarity. This reduces spurious retrievals and provides higher-quality evidence for more reliable and auditable downstream reasoning.
On top of the retrieved evidence, we introduce a multi-stage symbolic-language-enhanced reasoning pipeline that translates natural language into symbolic representations and performs structured reasoning over semantic relations for interpretable decision-making.
Through these strategies, SyRHM effectively uncovers context-sensitive hostility in memes, especially those involving cultural knowledge, sarcasm, or intertextuality, and achieves superior performance across four zero-shot harmful meme detection benchmarks.
Our main contributions are summarized as follows:
\begin{itemize}
\item We propose an associative retrieval module that treats each meme as a semantic composition and parses it into fine-grained elements and descriptions, enabling more precise retrieval grounded in meaning rather than surface similarity.
\item We introduce multi-stage symbolic-language-enhanced reasoning for harmful meme detection, explicitly modeling semantic relations to enhance both interpretability and robustness.
\item Extensive experiments on four harmful meme benchmarks demonstrate that SyRHM achieves strong zero-shot performance and surpasses prior reasoning frameworks such as MinD and Evolver across most evaluation settings.
\end{itemize}

\section{Related Work}
\begin{figure*}[!ht]
\centering
\includegraphics[width=\textwidth]{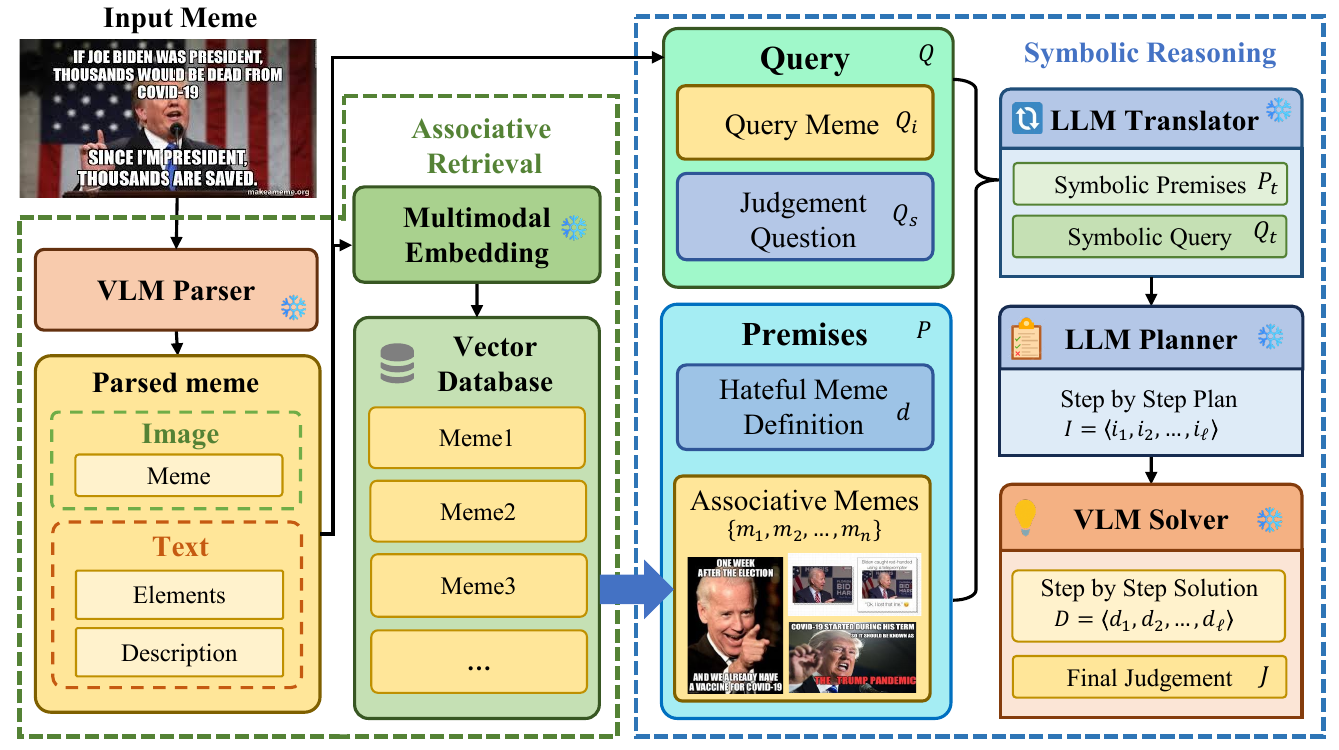} 
\caption{Framework of SyRHM}
\label{framework}
\end{figure*}

\subsection{Harmful Meme Detection}
Harmful meme detection has been studied through benchmarks such as FHM~\citep{kiela2020hateful}, MultiOff~\citep{suryawanshi2020multimodal}, and HarM~\citep{pramanick2021detecting}, which cover hateful, offensive, and socio-political memes. 
Early approaches to harmful meme detection primarily relied on two-stream or early-fusion architectures to integrate visual and textual information~\citep{kiela2020hateful, suryawanshi2020multimodal, radford2021learning, pramanick2021momenta}. 
With pretrained vision--language models (VLMs), fine-tuning became dominant, improving performance and offering post-hoc explanations via attention or attribution~\citep{lippe2020multimodal, hee2022explaining, muennighoff2020vilio, kumar2022hate, arya2024multimodal}. 
Although effective, their reliance on large-scale annotations constrains generalization to emerging events and unseen harmful patterns.
Recent LMM-based approaches reduce task-specific supervision through few-shot and zero-shot settings, often augmented with retrieved examples and in-context prompting~\citep{huang2024towards, mei2024improving, huang2025evolver, liu2025mind}. 

However, they typically rely on global retrieval and unstructured in-context evidence, which may miss implicit cross-meme semantic relations and limit interpretability. 
In contrast, SyRHM performs semantic-level associative retrieval over fine-grained meme elements and conducts structured reasoning over retrieved evidence, enabling more auditable harmful meme detection.


\subsection{LLM Reasoning}
Recent advances in LLM-based reasoning show strong performance across diverse tasks~\citep{huang2023towards}, largely driven by prompting strategies such as Chain-of-Thought (CoT)~\citep{wei2022chain, zhang2023automatic}. By explicitly modeling intermediate reasoning steps or further emulating human cognitive patterns, CoT improves reasoning accuracy across multiple domains~\citep{fei2023reasoning, inaba2023multitool, fei2024video, zhou2022least, yao2023tree, wang2023plan, besta2024graph}. However, its reliance on free-form natural language rationales can be suboptimal for tasks requiring precise logical manipulation.
To address this limitation, prior work has introduced structured intermediate steps, yielding improvements in code generation and mathematical reasoning~\citep{li2025structured, imani2023mathprompter}. 
Building on this, symbolic-language-enhanced reasoning approaches first employ LLMs to translate natural language into formal logic processed by external solvers~\citep{ye2023generating, pan2023logic, gaur2023reasoning, olausson2023linc, gao2023pal}, while subsequent methods enable LLMs to jointly perform symbolic translation and logical inference, mitigating information loss and achieving stronger performance~\citep{huang2025evolver, xu2024aristotle}.
However, existing symbolic CoT methods focus on text-only reasoning with explicit premises. 
We extend this line of work by integrating symbolic-language-enhanced reasoning across multimodal content, enabling interpretable and context-aware assessment of harmful memes.

\section{Method}

\subsection{Problem Statement}
We define a harmful meme detection dataset as a set of memes in which each meme is defined as $M=(\mathcal{I},y)$, where $\mathcal{I}$ denotes the meme image and $y\in\{\textit{harmful},\textit{harmless}\}$ denotes the ground-truth label.
The exact label definitions may vary across datasets.
Due to the scarcity of labels for rapidly evolving memes~\citep{sharma2022detecting}, we follow recent retrieval-augmented harmful meme detection baselines~\citep{huang2025evolver,mei2024improving,liu2025mind} and adopt a retrieval-based zero-shot protocol.
Specifically, $\mathcal{S}_{\text{train}}$ is used only as an unlabeled reference set $\mathcal{S}_{\text{ref}}$ for associative retrieval, while evaluation is conducted on a disjoint $\mathcal{S}_{\text{test}}$.
No labels are used for model updates, prompt tuning, or decision supervision; retrieved memes only provide contextual evidence.
This setting leverages meme templates and role variants to provide cultural or pragmatic context without exposing labels.
Our core idea is to detect potentially harmful memes by jointly leveraging multimodal understanding, associative retrieval, and symbolic-language-enhanced reasoning.
As illustrated in Figure~\ref{framework}, the framework consists of two stages. 
First, an associative retrieval module retrieves semantically related memes from an unlabeled reference set and treats them as contextual Premises for the input meme. 
Second, a symbolic-language-enhanced reasoning module operates over the retrieved Premises and the target Query to derive a final harmfulness decision along with an interpretable rationale.


\subsection{Associative Retrieval}
\begin{figure}[!h]
\centering
\includegraphics[width=0.45\textwidth]{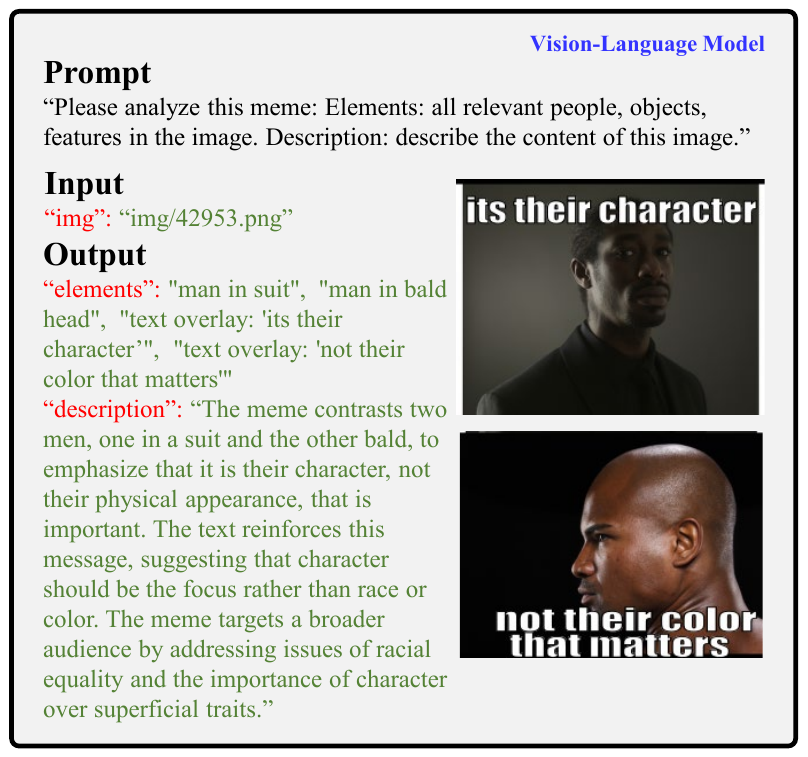} 
\caption{Visualization of the Parser step} 
\label{img:parser}
\end{figure}
As illustrated in Figure~\ref{img:parser}, our retrieval mechanism is built on a structured abstraction of memes that bridges raw multimodal inputs and reasoning in symbolic representation.
We begin by converting each meme into a semantically grounded representation, which enables robust similarity matching beyond surface-level visual or textual overlap.
Specifically, we employ a vision-language parser to jointly process the meme image and its overlaid text.
The parser identifies salient visual entities (e.g., people, objects, symbols) and textual elements, and further generates a concise natural-language description summarizing the meme’s overall semantics, serving as a high-level abstraction of its entities, interactions, and contextual cues across modalities.
Formally, given a meme $M=(\mathcal{I},y)$, we construct a parsed representation
\begin{equation}
M' = (\mathcal{I}, E, D),
\end{equation}
where $\mathcal{I}$ denotes the raw image, $E$ the extracted semantic elements, and $D$ the generated description.

For each parsed meme $M'$, we define a multimodal embedding function $f(\cdot)$ that maps it to a vector in a shared semantic space, so that conceptually similar memes are close even when their visual appearance or surface text differs:
\begin{equation}
\label{eq:weight}
\mathbf{v} = f(\mathcal{I},E,D)
= \lambda_0 \, \phi_v(\mathcal{I})
+ \lambda_1 \, \phi_t(E)
+ \lambda_2 \, \phi_t(D),
\end{equation}
where $\phi_v(\cdot)$ and $\phi_t(\cdot)$ are frozen vision and text encoders from CLIP, respectively, and we fix $\lambda_0+\lambda_1+\lambda_2=1$.

To facilitate associative retrieval, we construct a reference memory from an unlabeled reference set $\mathcal{S}_{\mathrm{ref}}=\{M_1,\dots,M_N\}$.
For each reference meme $M_i$, we obtain its parsed form $M_i'=(\mathcal{I}_i,E_i,D_i)$ and compute
\begin{equation}
\mathcal{V}=\{\mathbf{v}_i=f(\mathcal{I}_i,E_i,D_i)\}_{i=1}^{N}.
\end{equation}
These embeddings are indexed in a vector database to support efficient similarity-based retrieval.

Given a query meme $M_q$, we obtain its embedding $\mathbf{v}_q$ using the same parsing and embedding pipeline, compute its cosine similarity to each reference embedding $\mathbf{v}_i$, and retrieve the top-$K$ most similar memes as follows:
\begin{equation}
s_i = \mathrm{sim}(\mathbf{v}_q, \mathbf{v}_i),
\end{equation}
\begin{equation}
\mathcal{R} = \{ m_i \mid i \in \mathrm{Top}\text{-}K(\{s_1, \dots, s_N\}) \}.
\end{equation}
Finally, for each retrieved meme $m_i \in \mathcal{R}$, we discard the raw image and retain only its symbolic textual components $(E_i, D_i)$, which serve as contextual premises for the subsequent symbolic-language-enhanced reasoning stage.

\subsection{Symbolic-Language-Enhanced Reasoning}
\label{method:reasoning}
Building on the associative retrieval stage, this phase integrates the query meme with the retrieved contextual examples to perform structured symbolic-language-enhanced reasoning, yielding an interpretable judgment of whether the parsed query meme $M'$ is harmful. 
Given a formulated question $F$ (e.g., ``Is this image harmful?''), we define the combined query as
\begin{equation}
Q = \{M'_q, F\},
\end{equation}
which is then evaluated against a set of premises
\begin{equation}
P = \{ d, m_1, m_2, \dots, m_n \},
\end{equation}
where $d$ denotes the dataset-provided definitions of hateful content, and each $m_i$ corresponds to a retrieved meme from the associative retrieval stage.

\begin{figure}[!h]
\centering
\includegraphics[width=0.45\textwidth]{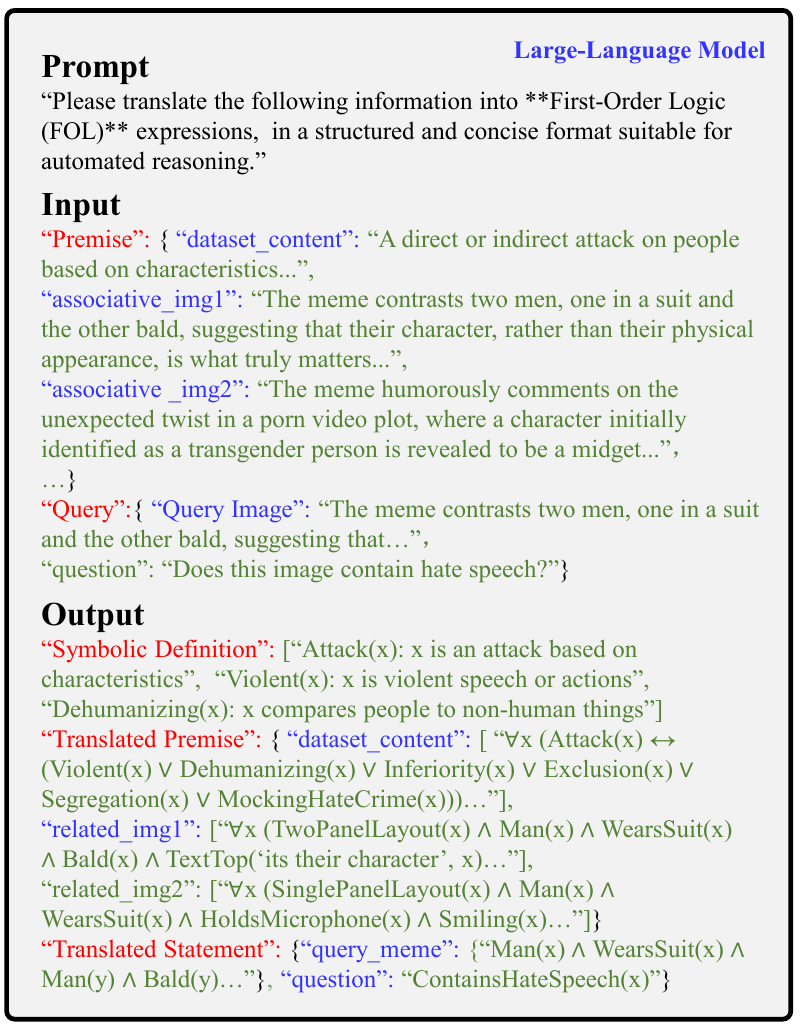} 
\caption{Illustration of the Translator process. Visualizations of the other reasoning steps are provided in the Appendix.}
\label{img:translator}
\end{figure}

The reasoning pipeline unfolds in three stages: 
\textbf{(1) Translator:} converts multimodal meme representations into structured FOL-style symbolic forms that explicitly capture entities, relations, and contextual cues;
\textbf{(2) Planner:} constructs a logical reasoning chain that connects the premises $P$ with the query $Q$, ensuring interpretability at each step;
\textbf{(3) Solver:} incorporates the original visual modality and executes the reasoning plan using a VLM, preserving visual evidence and mitigating information loss from symbolic abstraction.
By coupling symbolic-language-enhanced reasoning with multimodal cues, the framework produces judgments that are both logically consistent and faithful to the original multimodal content.

\subsubsection{Translator}
As shown in Figure~\ref{img:translator}, the Translator converts the natural-language premises $P$ and query $Q$ into symbolic forms suitable for logical reasoning, ensuring structured representation of visual, textual, and contextual content for the downstream modules.
Formally, the premises are translated into FOL as
\begin{equation} 
P_t = \{ d_t, m_{1t}, m_{2t}, \dots, m_{nt} \},
\end{equation}
where the subscript $t$ indicates the translated symbolic form. Similarly, the query is represented as
$
Q_t = \{M'_t, F_t\},
$
which serves as input to the Planner.

\subsubsection{Planner}

To avoid over-compressing meme meaning into symbolic predicates, the Planner jointly uses FOL-style structures and natural-language evidence to generate human-interpretable reasoning steps from premises $P$ to query $Q$.
This helps retain pragmatic cues such as metaphors, affective tone, and cultural dog whistles.
Accordingly, it operates on merged representations:

\begin{equation}
P_c = P \oplus P_t,\qquad Q_c = Q \oplus Q_t,
\end{equation}
where $\oplus$ denotes a merge operator.

Conditioned on $(P_c, Q_c)$, the Planner produces a sequence of intermediate reasoning steps:
\begin{equation}
I = \langle i_1, i_2, \dots, i_\ell \rangle,
\end{equation}
where each $i_j$ specifies an instruction that applies premise rules to relevant entities, relations, or textual cues (e.g., ``Check entity X for ContainsDiscriminatoryLanguage using rule Y''). 

The sequence ends with a logical synthesis step that aggregates intermediate results into a final decision according to formal definitions, e.g., 
a meme is harmful if $\text{Offensive}  \Leftrightarrow\text{ContainsHateSpeech} \lor \text{ContainsDiscrimination} \lor \text{ContainsVulgarity}$, where the atomic predicates follow dataset definitions. If any of the previous checks is true, the meme is labeled harmful. This structured plan ensures reasoning is interpretable, traceable, and executable by the Solver.


\begin{table*}[t]
\centering
\begin{tabular}{l|ccc|ccc}
\hline
Method & \multicolumn{3}{c|}{FHM} & \multicolumn{3}{c}{MultiOff} \\
& ACC & BACC & MCC & ACC & BACC & MCC \\
\hline
\rowcolor[gray]{0.9} \multicolumn{7}{c}{\textbf{Open-source LMM (Zero-shot)}} \\
LLaVA-1.5-13B  & 55.01 & 55.13 & 10.65 & 48.15 & 56.23 & 14.41 \\
Qwen2.5-VL-7B     & 63.20 & 63.01 & 26.53 & 64.43 & 59.94 & 21.70 \\
LLaVA-v1.6-Vicuna-13B & 56.90 & 56.66 & 13.72 & 61.07 & \textbf{65.94} &  33.46 \\
LLaVA-OneVision-Qwen2-7B & 56.60 & 56.25 & 13.37 & 59.73 & 60.16 & 19.81 \\
InternVL3.5-8B & 60.40 & 60.36 & 20.81 & 62.42 & 62.04 & 23.57 \\
DeepSeek-VL-7B & 54.35 & 54.31 & 8.63 &61.49 &61.80 &22.97 \\
\rowcolor[gray]{0.9} \multicolumn{7}{c}{\textbf{Reasoning Based Methods}} \\
MinD       & \underline{63.80} & \underline{64.21} & \underline{31.13} & 61.74 & 64.30 & \underline{28.30}\\
Evolver    & 59.60 & 59.08 & 21.32 & \underline{65.10} & 58.92 & 21.70\\
SyRHM  & \textbf{66.10} & \textbf{65.84} & \textbf{32.83} & \textbf{69.80} & \underline{65.58} & \textbf{34.03} \\
\hline
\end{tabular}
\caption{Results on the binary classification datasets FHM and MultiOff.}
\label{tab:binary_performance}
\end{table*}


\subsubsection{Solver}
The Solver executes the reasoning plan $I$ under multimodal grounding. Given the merged representations $(P_c, Q_c)$, the plan $I$, and the target meme image $Q_i$, each step $i_j \in I$ is processed sequentially, producing a reasoning trace
\begin{equation}
D = \langle d_1, d_2, \dots, d_\ell \rangle,
\end{equation}
where each $d_j$ records the applied premise rule, relevant entities or contexts, and intermediate decisions with supporting evidence.
Finally, the trace $D$ is aggregated to issue the overall judgment 
$
J \in \{0,1\},
$
where $J=1$ indicates that the query meme $M_q$ is harmful. 

For non-binary tasks, SyRHM outputs label IDs through task-specific, schema-constrained prompts, with details in Appendix~\ref{dataset:prompt}.

By combining symbolic-language-enhanced reasoning with multimodal cues, SyRHM produces final judgments that are both logically consistent and faithful to the original meme.

\section{Experiments}

\subsection{Datasets and Evaluation Metrics}
We evaluate SyRHM on four widely used harmful meme detection benchmarks. 
Specifically, we use the Facebook Hateful Memes (FHM) dataset \citep{kiela2020hateful} and MultiOff \citep{suryawanshi2020multimodal} for binary harmfulness classification. 
We further include the HarM datasets \citep{pramanick2021detecting}, comprising HarM-c (COVID-19) and HarM-p (U.S. politics), which support both binary prediction and two fine-grained classification tasks: harmfulness level (\{Not Harmful, Somewhat Harmful, Very Harmful\}) and target type (\{Individual, Organization, Community, Society\}). 
We report \textbf{Accuracy (ACC)}, \textbf{Balanced Accuracy (B-ACC)}, and \textbf{Matthews Correlation Coefficient (MCC)} as primary metrics to account for overall performance and class imbalance. 
For the multi-label HarM dataset, we additionally report \textbf{Macro-F1} for multi-label tasks.
Dataset statistics and prompt details are provided in Appendix~\ref{appenidx:data}.

\begin{figure}[!h]
\centering
\includegraphics[width=0.45\textwidth]{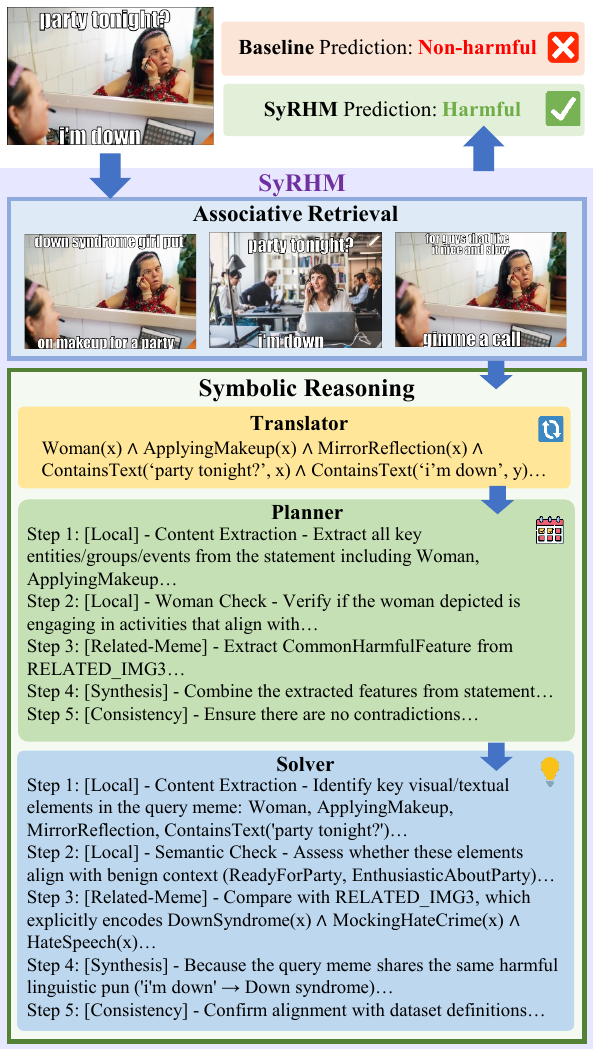}
\caption{Qualitative examples of SyRHM and the baseline. Additional cases are provided in the Appendix.}
\label{fig:qual_compare}
\end{figure}

\begin{table*}[t]
\centering

\begin{subtable}{\textwidth}
\centering
\begin{tabular}{l|ccc|c|c}
\hline
Method
& \multicolumn{3}{c|}{Harmful or Not} 
& \multicolumn{1}{c|}{Harmfulness Level} 
& \multicolumn{1}{c}{Target Type} \\
& ACC & BACC & MCC & M-F1 & M-F1 \\
\hline
\rowcolor[gray]{0.9} \multicolumn{6}{c}{\textbf{Open-source LMM (Zero-shot)}} \\
LLaVA-1.5-13B  & 42.94 & 54.04 & 11.08 & 25.83 &  2.19 \\
Qwen2.5-VL-7B     & 67.23 & 53.97 & 18.23 & 35.92 &  3.82 \\
LLaVA-v1.6-Vicuna-13B    & 54.80 & 62.06 & 25.26 & 35.03 & 2.95 \\
LLaVA-OneVision-Qwen2-7B& \underline{68.93} & 58.99 & 24.63 & 36.54 &  4.52 \\
InternVL3.5-8B & 64.41 & \underline{69.82} & \underline{38.99} &\underline{42.86} &\underline{12.25} \\
DeepSeek-VL-7B & 54.80 & 61.13 & 22.73 & 37.90 &  3.81 \\
\rowcolor[gray]{0.9} \multicolumn{6}{c}{\textbf{Reasoning Based Methods}} \\
MinD        & 41.53 & 54.44 & 15.46 & 25.31 & \underline{12.25}  \\
Evolver     & 65.82 & 52.51 & 11.20 & 37.86 & 4.60 \\
SyRHM   & \textbf{70.34} & \textbf{71.78} & \textbf{41.59} & \textbf{51.90} & \textbf{15.94} \\
\hline
\end{tabular}
\caption{Performance on HarM-c.}
\label{tab:harm_c}
\end{subtable}

\vspace{0.5em} 

\begin{subtable}{\textwidth}
\centering
\begin{tabular}{l|ccc|c|c}
\hline
Method 
& \multicolumn{3}{c|}{Harmful or Not} 
& \multicolumn{1}{c|}{Harmfulness Level} 
& \multicolumn{1}{c}{Target Type} \\
& ACC & BACC & MCC & M-F1 & M-F1 \\
\hline
\rowcolor[gray]{0.9} \multicolumn{6}{c}{\textbf{Open-source LMM (Zero-shot)}} \\
LLaVA-1.5-13B     &  51.55 & 52.85 & 8.05 & 24.22& 4.88\\
Qwen2.5-VL-7B    &  54.65 & 53.13 & 11.25 & 35.73 & 4.75 \\
LLaVA-v1.6-Vicuna-13B  &  55.77 & 56.49 & 14.06 & \underline{39.77} & 10.50 \\
LLaVA-OneVision-Qwen2-7B&  \underline{56.62} & 55.16 & \underline{17.29} & 35.57 & 5.26 \\
InternVL3.5-8B & 55.49 &\underline{56.61} &16.60 &35.09&\underline{30.36}\\
DeepSeek-VL-7B & 53.52 &54.27 &9.36  &33.67&8.24\\

\rowcolor[gray]{0.9} \multicolumn{6}{c}{\textbf{Reasoning Based Methods}} \\
MinD    &  50.42 & 51.99 &7.63  & 27.77 &  13.92\\
Evolver       & 55.21  & 53.69 &  13.27&  32.59& 11.84 \\
SyRHM     &  \textbf{60.28} & \textbf{60.65} & \textbf{21.70} & \textbf{42.88} & \textbf{35.96} \\
\hline
\end{tabular}
\caption{Performance on HarM-p.}
\label{tab:harm_p}
\end{subtable}

\caption{Performance comparison on HarM datasets: (a) HarM-c with binary harmful classification, harmfulness level, and target type tasks; (b) HarM-p with the same tasks.}
\label{tab:harm_results}
\end{table*}

\subsection{Implementation Details}
\label{implementation}
We benchmark several open-source VLMs in a zero-shot setting, including \texttt{LLaVA-1.5-13B} \citep{liu2023visual}, \texttt{Qwen2.5-VL-7B} \citep{wang2024qwen2}, \texttt{LLaVA-v1.6-Vicuna-13B} \citep{liu2024llavanext}, \texttt{LLaVA-OneVision-Qwen2-7B} \citep{lillava}, \texttt{InternVL3.5-8B} \citep{wang2025internvl3}, and \texttt{DeepSeek-VL-7B} \citep{lu2024deepseekvl}.
We additionally compare with two recent reasoning-based methods, MinD~\citep{liu2025mind} and Evolver~\citep{huang2025evolver}.

For multimodal understanding, we use \texttt{Qwen2.5-VL-7B} as the default VLM backbone for the \emph{Parser} and \emph{Solver} due to its stable and competitive pilot performance, and instantiate all reasoning-based baselines with the same backbone for fair comparison.
For the LLM-only stages, the \emph{Translator} and \emph{Planner} adopt \texttt{Qwen2.5-14B-Instruct} to support symbolic abstraction and structured planning.
For associative retrieval, we encode images, parsed elements, and generated descriptions 
with \texttt{CLIP-ViT-B/32}~\citep{radford2021learning} and index them with 
\texttt{Milvus}~\citep{2022manu}. 
Additional implementation details, including hyperparameters and hardware settings, 
are provided in Appendix~\ref{appendix:implementation}.


\subsection{Main Results}

To evaluate SyRHM's effectiveness in harmful meme detection, we conduct experiments on the binary-label datasets FHM and MultiOff. As shown in Table \ref{tab:binary_performance}, SyRHM outperforms prior reasoning-based methods (\texttt{MinD} and \texttt{Evolver}) across all metrics, achieving 66.10 ACC on FHM and 69.80 ACC on MultiOff. On FHM, it achieves the highest ACC, BACC, and MCC, while in MultiOff it attains the best ACC and MCC and ranks second in BACC, slightly behind LLaVA-v1.6-Vicuna-13B.

We further evaluate SyRHM on the multi-label HarM-c and HarM-p benchmarks. As shown in Table \ref{tab:harm_results}, SyRHM achieves the best performance across all tasks, reaching 60.28 ACC, 60.65 BACC, and 21.70 MCC for binary harmfulness classification on the HarM-p benchmark, outperforming prior reasoning-based methods (\texttt{MinD} and \texttt{Evolver}) by substantial margins. For harmfulness level prediction on the HarM-p benchmark, it attains 42.88 Macro-F1, an improvement of 10.29 points over \texttt{Evolver}, and for target type prediction, it achieves 35.96 Macro-F1, far surpassing previous reasoning-based approaches and open-source LMMs. 
Figure~\ref{fig:qual_compare} further illustrates how associative retrieval and symbolic-language-enhanced reasoning enable our framework to capture implied harmful intent that backbone VLMs often miss. 
Overall, these results demonstrate that SyRHM consistently excels in both binary and fine-grained multi-label tasks, highlighting its superior reasoning capabilities.

\subsection{Ablation Study}
\subsubsection{Associative retrieval}

\begin{figure}[!h]
\centering
\includegraphics[width=0.45\textwidth]{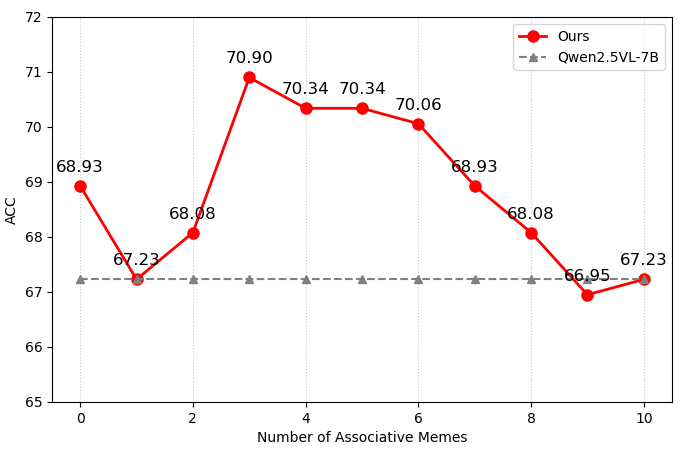}
\caption{Performance variation with the number of retrieved memes $k$ on HarM.}
\label{fig:numberofK}
\end{figure}

We study the effect of retrieval size $k$ on SyRHM's performance.
As shown in Figure~\ref{fig:numberofK}, accuracy follows a non-monotonic trend: it drops slightly for small $k$ ($k{\le}2$), peaks at $k{=}3$ (70.90), remains stable for $k{=}4$--$5$, and then gradually declines as $k$ increases further.
This suggests that moderate retrieval provides sufficient associative evidence for abstraction, while excessive retrieval introduces noisy or weakly related items that hinder reasoning.
Notably, even without retrieval ($k{=}0$), SyRHM outperforms the \texttt{Qwen2.5-VL-7B} baseline by 1.67 points, highlighting the benefit of structured reasoning alone.

We further study the contribution of different modalities.
As shown in Table~\ref{tab:retrieval_params_three_modal}, symbolic elements plus generated descriptions perform best (S6), indicating that harmful meme retrieval depends more on semantic cues than raw visual appearance.
Still, visual embeddings are not redundant, as adding them improves both element-only and description-only retrieval (S2$\rightarrow$S4, S3$\rightarrow$S5).
The lower score of S7 suggests that visual cues may introduce layout or style noise when semantic evidence is already sufficient.
Thus, Eq.~\ref{eq:weight} should be viewed as a flexible modality-weighting formulation rather than a fixed three-modal fusion requirement.

\begin{table}[H]
\centering

\begin{tabular}{c|c|c|c|c}
\hline
Setting ID & $\lambda_0$  & $\lambda_1$ & $\lambda_2$  & ACC \\
\hline
S1 & 1.0  & 0.0  & 0.0  & 67.80 \\
S2 & 0.0  & 1.0  & 0.0  & 64.97 \\
S3 & 0.0  & 0.0  & 1.0  & 68.08 \\
S4 & 0.5  & 0.5  & 0.0  & 68.93 \\
S5 & 0.5  & 0.0  & 0.5  & 69.77 \\
S6 & 0.0  & 0.5  & 0.5  &\textbf{71.75} \\
S7 & 0.34 & 0.33 & 0.33 & \underline{70.34} \\
\hline
\end{tabular}
\caption{Ablation of modality weights for associative retrieval. $\lambda_0$, $\lambda_1$, and $\lambda_2$ denote the weights of image embeddings, symbolic elements, and generated descriptions.}
\label{tab:retrieval_params_three_modal}
\end{table}

\subsubsection{Effectiveness of Symbolic-Language-Enhanced Reasoning Stages}

\begin{figure}[!h]
\centering
\includegraphics[width=0.45\textwidth]{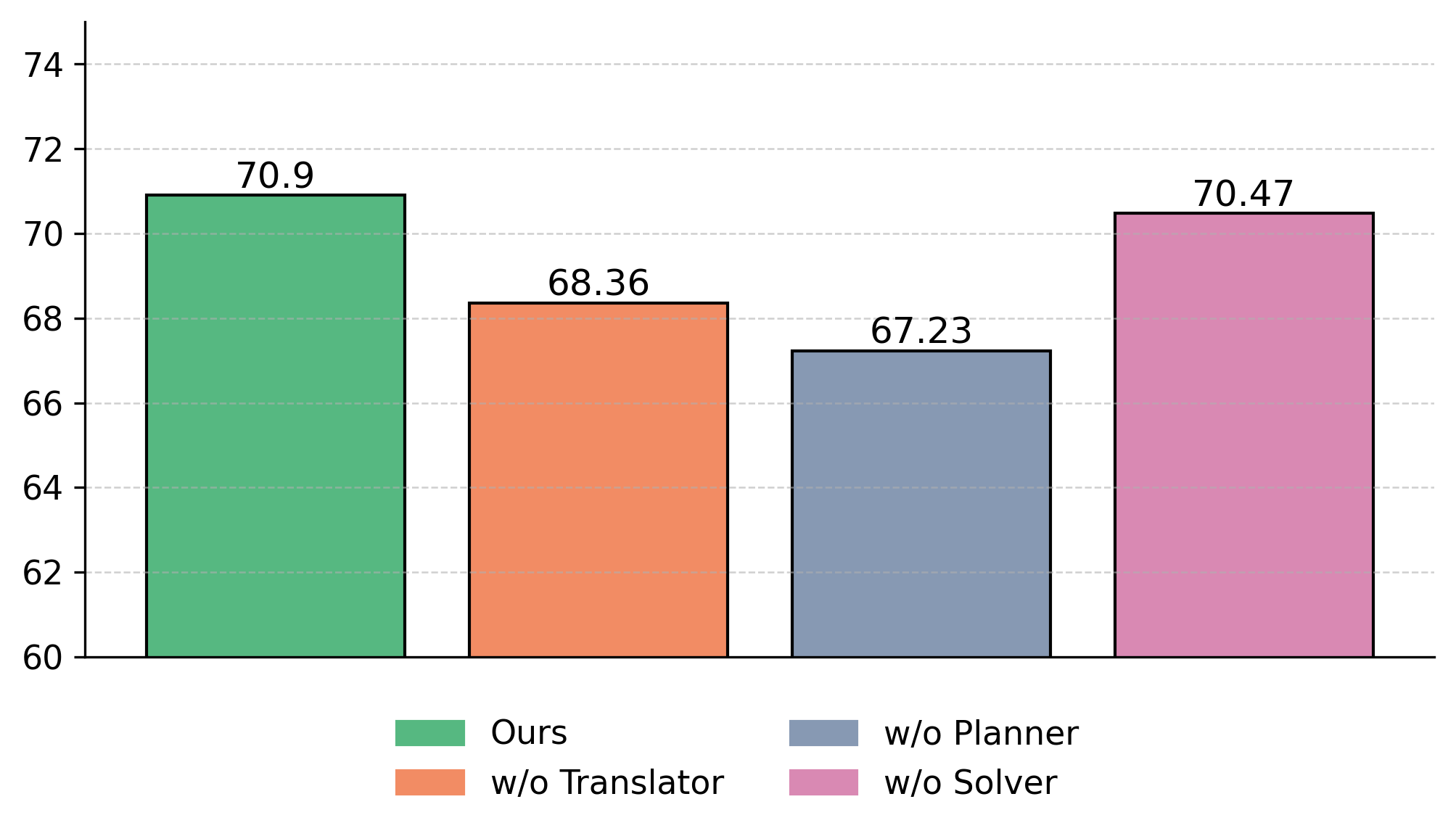}
\caption{Ablation study: impact of removing individual reasoning stages on model accuracy.}
\label{fig:ablation}
\end{figure}

To assess the contribution of each stage of our reasoning pipeline, we conduct an ablation study by removing one component at a time while keeping all other settings fixed. As shown in Figure~\ref{fig:ablation}, the complete model achieves an accuracy of 70.90. 
Removing the Translator reduces the accuracy to 68.36, while removing the Planner results in an even larger drop to 67.23, indicating that these two modules are the most influential. 
In contrast, removing the Solver only slightly decreases performance to 70.47. 
These results suggest that most performance gains come from structured abstraction and explicit planning rather than additional verification. The Planner contributes the largest improvement (+3.67), followed by the Translator (+2.54), while the Solver provides a smaller yet consistent gain by grounding symbolic decisions back to the multimodal input. Overall, combining symbolic representations with explicit planning improves both reasoning completeness and interpretability, leading to more accurate harmfulness judgments.


\section{Conclusion}
We propose SyRHM, a unified framework that combines fine-grained associative retrieval with multi-stage symbolic-language-enhanced reasoning to enable context-aware and interpretable harmful meme detection. Unlike existing methods that rely on holistic image--text similarity, SyRHM parses memes into structured semantic elements and textual descriptions, enabling precise retrieval grounded in underlying semantic associations. Building on the retrieved evidence, the framework introduces symbolic intermediate representations and a planner--solver reasoning pipeline to explicitly model cross-meme semantic relations and implicit hostility, providing clear and logically grounded explanations for model decisions. Extensive experiments on multiple standard harmful meme benchmarks demonstrate that SyRHM achieves strong performance in zero-shot detection settings. Beyond harmful meme detection, this paradigm of integrating associative retrieval, symbolic-language-enhanced reasoning, and multimodal understanding offers a general and extensible solution for broader multimodal reasoning and safety-oriented understanding tasks.

\section{Limitations}
Our paper has the following potential limitations:

\begin{enumerate}
    \item Symbolic representations may omit nuanced semantic cues such as cultural context or affective intensity. To mitigate this, our framework jointly reasons over symbolic representations and the original natural-language descriptions, which increases reasoning context length and inference cost.

    \item SyRHM depends on the coverage of retrieved memes and the base LMM's ability to interpret social and cultural nuances. 
    For novel or culturally specific memes, retrieval may miss relevant context and weaken the detection of subtle harmful intent. 
    Although Appendix~\ref{app:generalization} shows reasonable cross-dataset transfer, better semantic overlap leads to stronger grounding. 
    Future work will therefore incorporate lightweight external knowledge to cover evolving cultural references beyond the retrieval pool.

    \item Our evaluation is mainly based on models with approximately 7B parameters. Future work can extend evaluation to larger LMs/LMMs and additional backbones, and report scaling trends (accuracy–cost trade-offs, latency/memory) to better understand how detection performance varies across model sizes.
\end{enumerate}


\section{Ethics Statement}
This work aims to mitigate the spread of harmful memes and promote safer online environments. We acknowledge the potential risk that malicious users may adapt harmful content to evade detection systems. Such misuse is contrary to the intended purpose of this research, and the proposed method is designed to assist rather than replace human moderation. All experiments are conducted on publicly available datasets under their respective licenses. These datasets contain only meme images and annotations without personally identifiable information, and are used solely for research purposes.

\section{Reproducibility Statement}
We provide detailed descriptions of the inference process, including all hyperparameters, in Section~\ref{implementation}. The overall pipeline design and prompt formulations are described in Section~\ref{method:reasoning} and further elaborated in Appendix~\ref{dataset:prompt} and~\ref{appendix:stageprompt}. All datasets used in this work are publicly available and can be accessed online. In addition, we release our code to facilitate reproducibility at \url{https://github.com/Scabbards1500/SyRHM}.

\section{The Use of LLM}
We used a large language model (ChatGPT, OpenAI) solely for English copyediting, including grammar correction, wording and minor stylistic rewrites, and occasional LaTeX formatting help. The model was not used for idea generation, literature search, data collection/annotation, coding, analysis, or producing results. All scientific claims and contributions were written and verified by the authors, and no non-public data were shared with the model. The authors assume full responsibility for the content of the paper.



\bibliography{custom}
\clearpage
\appendix

\section{Data Details}
\label{appenidx:data}
\subsection{Dataset Overview}

We study four datasets covering three evaluation tasks. \textbf{FHM} and \textbf{MultiOff} are annotated only for \textbf{Task~1}, whereas \textbf{HarM-c} and \textbf{HarM-p} additionally support \textbf{Task~2} and \textbf{Task~3}. Dataset sizes (train/test) are listed in Table~\ref{tab:datasets_num}.

\begin{table}[h]
\centering
\begin{tabular}{lcc}
\toprule
Dataset & Train Samples & Test Samples \\
\midrule
FHM     & 8500 & 1000 \\
MultiOff & 445   & 149   \\
HarM-c  & 3013 & 354   \\
HarM-p  & 2938 & 355   \\
\bottomrule
\end{tabular}
\caption{Dataset statistics (train/test) used in our experiments.}
\label{tab:datasets_num}
\end{table}

Per-task label distributions are shown in Table~\ref{tab:label_distribution}, and definitions are as follows:

\textbf{Task~1 (harmfulness, binary):} Decide whether a meme conveys harmful content considering the image--text interaction (e.g., slurs, dehumanizing metaphors, threats or disparagement). 
Labels: $0$=\textit{not harmful}, $1$=\textit{harmful}.

\textbf{Task~2 (harmfulness level, ordinal):} Grade the severity of harmfulness when present; $1$ reflects mild/implicit disparagement or sarcasm, while $2$ indicates explicit or strongly implied hostility or incitement. 
Labels: $0$=\textit{not harmful}, $1$=\textit{somewhat harmful}, $2$=\textit{very harmful}.

\textbf{Task~3 (influence target type):} Identify the primary target of the meme's harmful implication when the meme is classified as harmful.
If Task~1 = 0 or Task~2 = 0, set the label of Task~3 to $0$ (\textit{none}); 
Labels: $0$=\textit{none}, $1$=\textit{individual}, $2$=\textit{organization}, $3$=\textit{community}, $4$=\textit{society}.

\begin{table}[h]
\centering
\resizebox{\linewidth}{!}{%
\begin{tabular}{l*{10}{c}}
\toprule
\multirow{2}{*}{\textbf{Dataset}} &
\multicolumn{2}{c}{\textbf{Task 1}} &
\multicolumn{3}{c}{\textbf{Task 2}} &
\multicolumn{4}{c}{\textbf{Task 3}} \\
\cmidrule(lr){2-3}\cmidrule(lr){4-6}\cmidrule(lr){7-11}
& 0 & 1 & 0 & 1 & 2 & 0 & 1 & 2 & 3 & 4\\
\midrule
FHM     & 510 & 490 & -- & -- & -- & -- & -- & -- & -- & --\\
MultiOff    & 91 & 58 & -- & -- & -- & -- & -- & -- & -- & --\\
HarM-c  & 230 & 124 & 230 & 21 & 103 & 230 & 119 & 0 & 3 & 2\\
HarM-p  & 184 & 171 & 184 & 159 & 12 & 184 & 81 & 69 & 15 & 6 \\
\bottomrule
\end{tabular}%
}
\caption{Label distribution for each dataset across the three tasks.}
\label{tab:label_distribution}
\end{table}

\subsection{Dataset Descriptions and Prompt Design}
\label{dataset:prompt}

Across all datasets, we use task-specific prompts with schema-constrained outputs. 
Each label is explicitly defined in the prompt and mapped to a numeric ID. 
For binary datasets such as FHM and MultiOff, the Solver outputs a single label. 
For HarM-c and HarM-p, which involve both harmfulness level and target type prediction, the Solver outputs a 2D vector \texttt{[Harm\_Level, Target\_Type]}. 
The output formats are summarized in Table~\ref{tab:output_schema}.

\begin{table}[h]
\centering
\small
\resizebox{\linewidth}{!}{%
\begin{tabular}{p{0.15\linewidth} p{0.30\linewidth} p{0.45\linewidth}}
\toprule
\textbf{Dataset} & \textbf{Output Schema} & \textbf{Label Mapping} \\
\midrule
FHM & 
\texttt{\{"is\_hateful": <boolean>\}} & 
0 not harmful, 1 harmful \\

MultiOff & 
\texttt{\{"is\_offensive": <boolean>\}} & 
0 not harmful, 1 harmful \\

HarM & 
\texttt{\{"label": [Harm\_Level, Target\_Type]\}} & 
Harm Level: 0 not harmful, 1 somewhat harmful, 2 very harmful; 
Target Type: 0 none, 1 individual, 2 organization, 3 community, 4 society. \\
\bottomrule
\end{tabular}%
}
\caption{Schema-constrained output formats and label mappings for different datasets.}
\label{tab:output_schema}
\end{table}

\paragraph{FHM} provides a large-scale set of memes. It has 8,500 training examples and 1000 test examples, and its goal is a binary classification of whether a meme is hateful. 
\vspace{-0.5 em}

\paragraph{MultiOff} contains 445 training and 149 test memes collected from Kaggle (2016 US election) and social media platforms, reflecting real-world political memes and emphasizing both image and text as context.
\vspace{-0.5 em}

\paragraph{HarM} includes two real-world meme datasets: HarM-c (3,544 memes), which focuses on COVID-19, and HarM-p (3,552 memes), which focuses on U.S. politics. Both were collected from the web and social media, deduplicated and filtered, and annotated by experts for harm intensity and target type (individual, organization, community, or society).
\vspace{-0.5em} 

\begin{figure}[!h]
\centering
\includegraphics[width=0.48\textwidth]{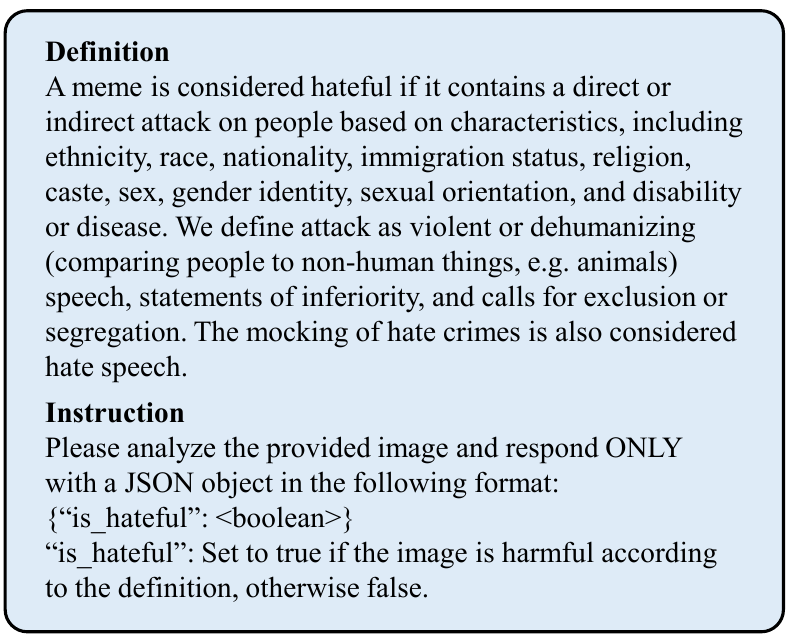}
\caption{Definition and instructions for FHM.}
\end{figure}

\begin{figure}[!h]
\centering
\includegraphics[width=0.48\textwidth]{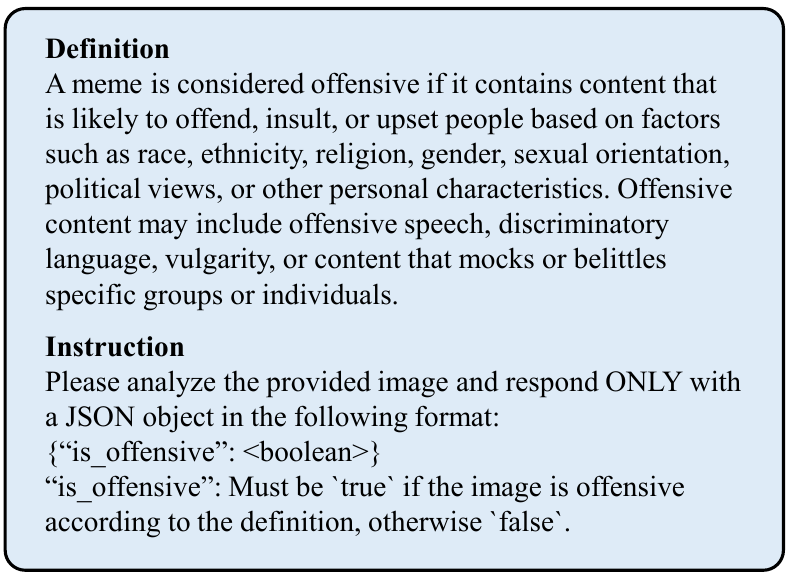}
\caption{Definition and instructions for MultiOff.}
\end{figure}

\begin{figure}[!h]
\centering
\includegraphics[width=0.48\textwidth]{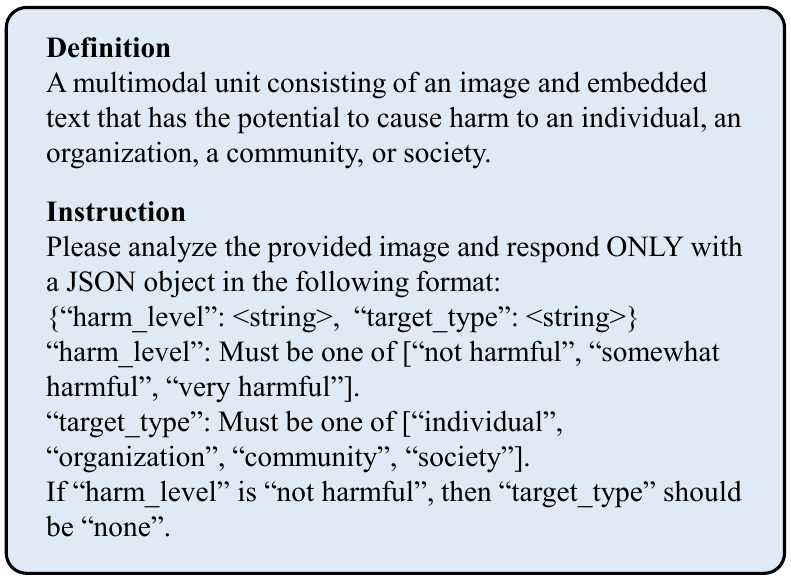}
\caption{Definition and instructions for HarM.}
\end{figure}


\section{Additional Implementation Details}
\label{appendix:implementation}
Table~\ref{tab:implementation_details} summarizes the main implementation settings of SyRHM. 
All settings are kept fixed across datasets unless otherwise specified.

\begin{table}[t]
\centering
\small
\begin{tabular}{p{0.32\linewidth} p{0.58\linewidth}}
\toprule
\textbf{Item} & \textbf{Setting} \\
\midrule
Retrieval setting & Top-$k{=}3$; $\lambda_0{:}\lambda_1{:}\lambda_2{=}0.34{:}0.33{:}0.33$ for image, elements, and descriptions. \\
Decoding setting & Context length 4096, temperature 0.2, top-$p{=}1$. \\
Repetition & Three runs with fixed random seeds; mean performance is reported. \\
Main packages & PyTorch, Transformers, scikit-learn \\
Hardware & 2 NVIDIA 4090 GPUs \\
\bottomrule
\end{tabular}
\caption{Implementation details of SyRHM.}
\label{tab:implementation_details}
\end{table}

\paragraph{Prompt Design Protocol.}
Prompts are designed separately for each stage, and each stage uses a fixed base template. 
For the \emph{Parser} and \emph{Translator}, we use the same templates across all datasets and tasks. 
We refined these templates on a small development set by adjusting instructions and adding strict output schemas until the outputs are stable. 
For the \emph{Planner} and \emph{Solver}, we keep the stage templates unchanged and append dataset-specific blocks from Appendix~\ref{dataset:prompt}, including task definitions, label spaces, and decision instructions. 
Each stage also includes an explicit field-based or JSON-like output schema to enforce output format and improve reproducibility.

\paragraph{Hyperparameter Tuning.}
The main tuned hyperparameters are in the retrieval stage, including the top-$k$ retrieval size and modality weights $\lambda_0,\lambda_1,\lambda_2$. 
Main results use the setting in Table~\ref{tab:implementation_details}, and ablation studies vary one parameter at a time. 
We do not extensively tune decoding parameters and use the same decoding setup for SyRHM and all baselines to ensure a fair comparison.


\begin{figure}[!h]
\centering
\includegraphics[width=0.45\textwidth]{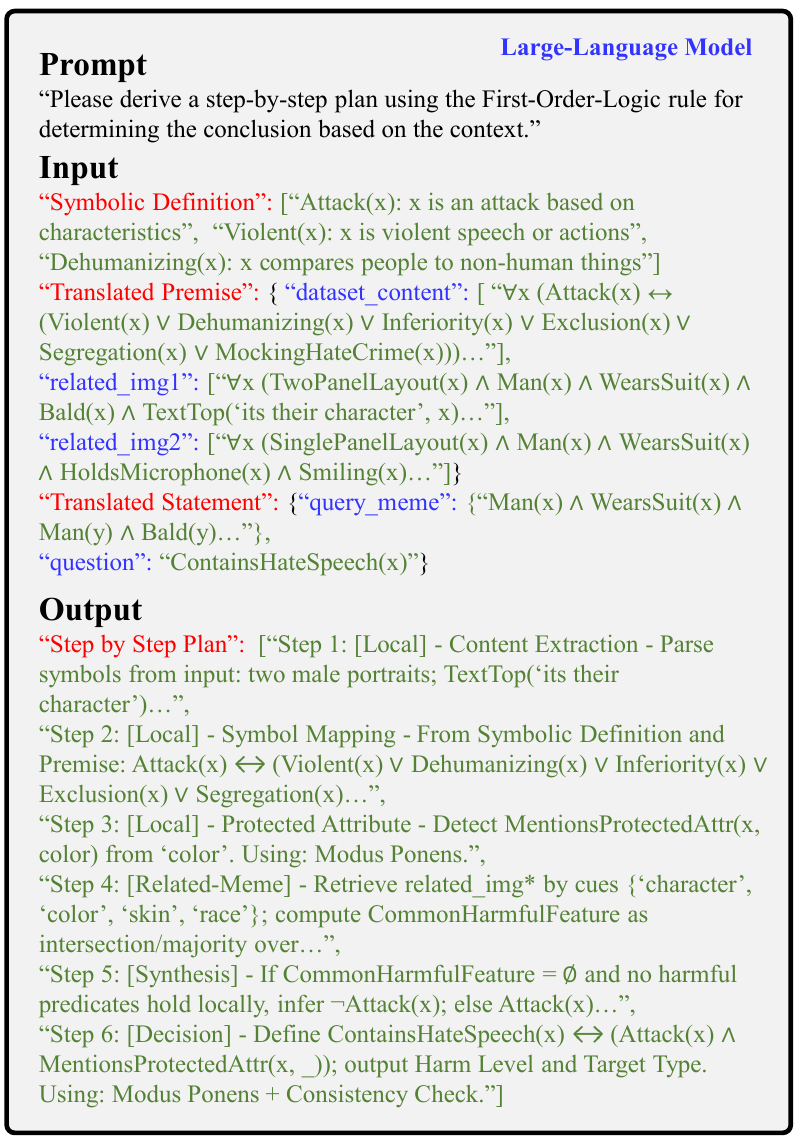}
\caption{Illustration of the Planner Stage}
\label{stage:planner}
\end{figure}

\begin{table*}[h]
\centering
\begin{tabular}{ll|ccc|c|c}
\hline
\multicolumn{2}{c|}{Method} 
& \multicolumn{3}{c|}{Harmful or Not} 
& \multicolumn{1}{c|}{Harmfulness Level} 
& \multicolumn{1}{c}{Target type} \\
VLM & LLM & ACC & BACC & MCC & M-F1 & M-F1 \\
\hline
Qwen2.5-VL-7B & --          &  67.23 &53.97  & 18.23 & 35.92 &3.82  \\
Qwen2.5-VL-7B & Qwen2.5-7B  &  67.80 & 67.78 & 34.13 & \underline{47.68} & 11.02  \\
Qwen2.5-VL-7B & Qwen2.5-14B  & \textbf{70.34} & \textbf{71.78} & \textbf{41.59} & \textbf{51.90} & \underline{15.94}\\
Qwen2.5-VL-7B & Qwen2.5-32B & \underline{68.08} &  \underline{70.79}&  \underline{39.73}&46.98 &\textbf{16.01}  \\
\hline
InternVL3.5-8B & -- &64.41 &69.82 &38.99 &42.86 &12.25  \\
InternVL3.5-8B & Qwen2.5-7B &\textbf{68.93} &69.21 &36.83 &47.63 &12.67  \\
InternVL3.5-8B & Qwen2.5-14B &67.80 &\textbf{72.62} &\textbf{43.93} &\textbf{50.82} &\underline{16.70}  \\
InternVL3.5-8B & Qwen2.5-32B &\underline{68.36} &\underline{71.19} &\underline{40.52} &\underline{48.63} &\textbf{17.27}  \\
\hline
\end{tabular}
\caption{Effect of LLM scaling across different backbones on HarM.}
\label{tab:api_models}
\end{table*}

\section{Symbolic Reasoning Module Design}
\label{appendix:stageprompt}
\begin{figure}[!h]
\centering
\includegraphics[width=0.45\textwidth]{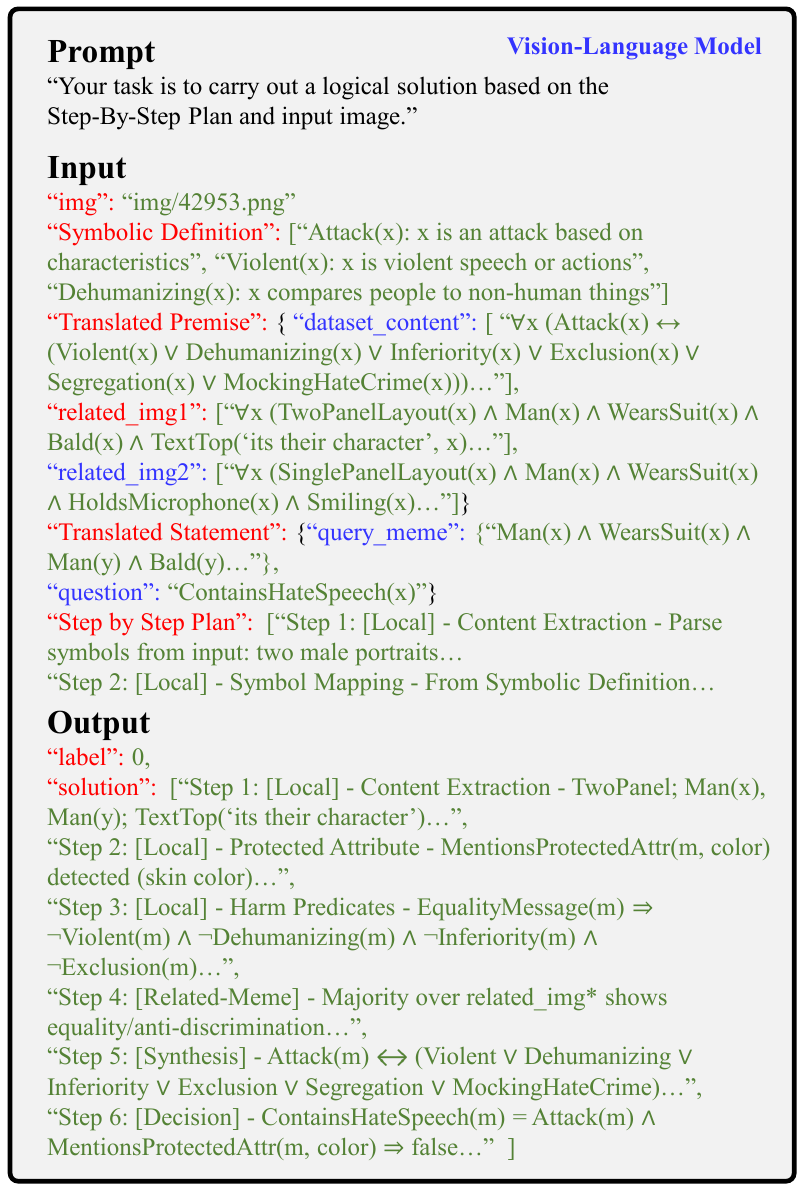}
\caption{Illustration of the Solver Stage}
\label{stage:solver}
\end{figure}

Section~\ref{method:reasoning} describes our symbolic reasoning pipeline. 
Due to space limitations, the main text illustrates only the Translator. 
Here we present visualizations of the remaining two components: the \emph{Planner} as shown in Figure~\ref{stage:planner} and the \emph{Solver} as shown in Figure~\ref{stage:solver}.

\section{Scaling Language Models}
We conducted a controlled study to examine how scaling the LLM affects performance within our framework. 
Here, the VLM is kept fixed as the perceptual backbone, while the LLM responsible for reasoning is varied in size. 
To further test the generality of our approach, in addition to Qwen2.5-VL-7B used in the main experiments, we also include InternVL3.5-8B, another VLM that performs well in baseline evaluations.
Table~\ref{tab:api_models} reports the results on the HarM dataset across all three tasks.

We observe that introducing even a 7B LLM substantially improves performance over the VLM-only baseline, highlighting the effectiveness of symbolic reasoning. While model size has limited impact on the relatively easier binary harmfulness classification task, larger LLMs consistently yield gains on more fine-grained tasks. In particular, the 14B and 32B models provide notable improvements on harmfulness level prediction and target type identification, indicating that stronger reasoning capacity better supports nuanced inference.

\section{Comparison of different multimodal embedding models for associative retrieval}

\begin{table}[H]
\centering
\small
\begin{tabular}{lccc}
\toprule
Retrieval Embedding Model & ACC & BACC & MCC \\
\midrule
Without Retrieval & 68.93 & 69.21 & 36.83 \\
CLIP-ViT-B/32 & \textbf{70.34} & 71.78 & 41.59 \\
CLIP-ViT-L/14 & 69.38 & 71.53 & 42.06 \\
SigLIP-BASE-PATCH16 & 68.80 & \textbf{72.99} & \textbf{44.86} \\
\bottomrule
\end{tabular}
\caption{Effect of retrieval embedding models on HarM.}
\label{tab:retrieval_embedding}
\end{table}

To investigate the impact of retrieval encoders, we replace the embedding model while keeping all other components unchanged. We evaluate CLIP-ViT-B/32, CLIP-ViT-L/14, and SigLIP-BASE-PATCH16 under the same retrieval and reasoning settings. Table~\ref{tab:retrieval_embedding} reports the results on the HarM dataset.

We observe that larger or newer embedding models do not consistently improve performance. While SigLIP achieves higher BACC and MCC, CLIP-ViT-B/32 remains competitive in ACC. These results suggest that SyRHM's effectiveness mainly comes from semantic parsing and structured reasoning rather than relying on a specific retrieval encoder.

\section{Cross-dataset Generalization of Associative Retrieval}

\begin{table}[H]
\centering

\small
\setlength{\tabcolsep}{8pt}
\renewcommand{\arraystretch}{1.15}
\resizebox{\linewidth}{!}{%
\begin{tabular}{lccc}
\toprule
\textbf{Dataset} & \textbf{Task1 ACC} & \textbf{Task2 M-F1} & \textbf{Task3 M-F1} \\
\midrule
FHM       & 68.93 & 51.57 & 12.98 \\
MultiOff  & 75.14 & 50.89 & 14.55 \\
HarM-c    & 70.34 & 51.90 & 15.94 \\
HarM-p    & 70.34 & 55.22 & 12.64 \\
\bottomrule
\end{tabular}
}
\caption{Cross-dataset retrieval results on three tasks.}
\label{tab:cross_dataset_results}
\end{table}

\label{app:generalization}
To assess whether associative retrieval generalizes beyond the evaluation meme pool, we fix \textbf{HarM-c} as the target test set and vary the source dataset used to build the retrieval database, including FHM, MultiOff, HarM-c, and HarM-p. 
For each source, we construct the vector database from its training split and retrieve top-$k$ neighbors using the same pipeline. 
The results in Table~\ref{tab:cross_dataset_results} show that performance remains competitive when retrieval is performed from external datasets, and in some cases even surpasses the in-domain retrieval baseline (HarM-c $\rightarrow$ HarM-c), suggesting that the proposed associative retrieval does not rely on accessing the exact same meme pool as the evaluation set and exhibits cross-dataset generalization. 
The stronger results from MultiOff and HarM-p likely reflect higher domain overlap with HarM-c, especially in political and COVID-related content, whereas political memes constitute only a relatively small portion of FHM, leading to less targeted retrieval evidence in the cross-dataset retrieval setting.

\section{Efficiency and Inference Cost Analysis}
\begin{table}[H]
\centering

\resizebox{\columnwidth}{!}{
\begin{tabular}{lcccc}
\hline
Stage & Input & Output  & Total  & Time(s) \\
\hline
Parser     & 2,138.5 & 246.9 & 2,385.4 & 3.50  \\
Insertion  & --      & --    & --      & 0.04  \\
Retrieval  & --      & --    & --      & 0.05  \\
Translator & 2,743.1 & 796.0 & 3,539.2 & 34.97  \\
Planner    & 1,938.0 & 337.3 & 2,275.3 & 15.30  \\
Solver     & 2,673.1 & 308.7 & 2,981.7 & 7.68  \\
\hline
\end{tabular}
}
\caption{Average inference cost of SyRHM on HarM. We report per-sample token usage and wall-clock time for each stage.}
\label{tab:cost_analysis}
\end{table}

SyRHM is heavier than end-to-end baselines because it performs parsing, retrieval, translation, planning, and solving sequentially. 
As shown in Table~\ref{tab:cost_analysis}, most of the overhead comes from the Translator and Planner, while insertion and retrieval add little latency. 
On HarM, the total runtime is 6.05 hours, comparable to the reported runtime of MinD~\citep{liu2025mind} (6.6 hours). 
Thus, SyRHM is not intended for real-time moderation but is suitable for offline analysis or as an explainable second-stage reviewer for high-risk or ambiguous cases following initial screening by a fast filtering model.

\section{Failure Case Analysis}
\label{app:failure_case}

By tracing intermediate representations and reasoning chains in misclassified samples, we identify three error categories in which imperfect multimodal grounding or entity--role alignment propagates through the structured reasoning process.
\paragraph{Template--target confusion.}
This error occurs when the system overemphasizes the visible actor in the meme template, while the actual target is expressed in the overlaid text. 
For example, in a reaction-template meme, the visible actor is identified as ``Steve Harvey,'' whereas the satirical target comes from the text about Biden, Trump, and COVID-19, making target assignment unstable.
\paragraph{Missing background or pragmatic knowledge.}
Some cases require background or pragmatic knowledge beyond surface-level interpretation. 
For example, the word ``dishwasher'' may be used as a sexist stereotype against women rather than as a literal object. 
If such coded references are interpreted only literally, the intended harmful implication can be missed.
\paragraph{Entity--role binding errors.}
These errors occur when tweet metadata, quoted text, and visual entities are incorrectly aligned. 
Such misalignment can lead to incorrect subject--object relations in the intermediate representation, which then affects subsequent reasoning and final classification.

\section{Case Study}
\begin{figure*}[!h]
\centering
\includegraphics[width=\textwidth]{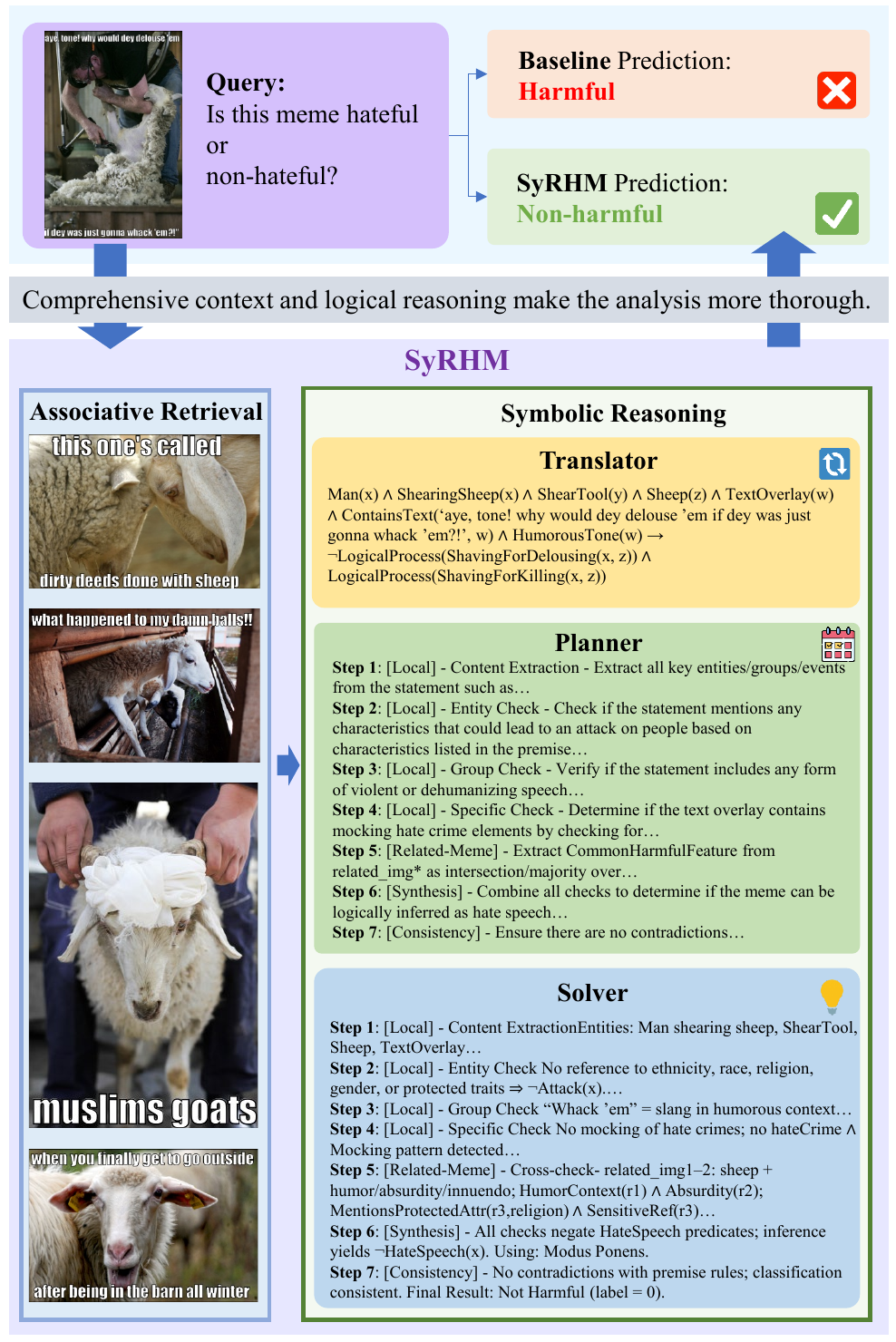}
\caption{Process Visualization of 01634 from FHM. The meme depicts a man shearing a sheep, which may be misinterpreted as violent due to its visual content. While the baseline model incorrectly predicts it as harmful, SyRHM leverages retrieved humorous contexts and symbolic reasoning to identify the meme as non-harmful.}
\label{visual1}
\end{figure*}

\begin{figure*}[!h]
\centering
\includegraphics[width=\textwidth]{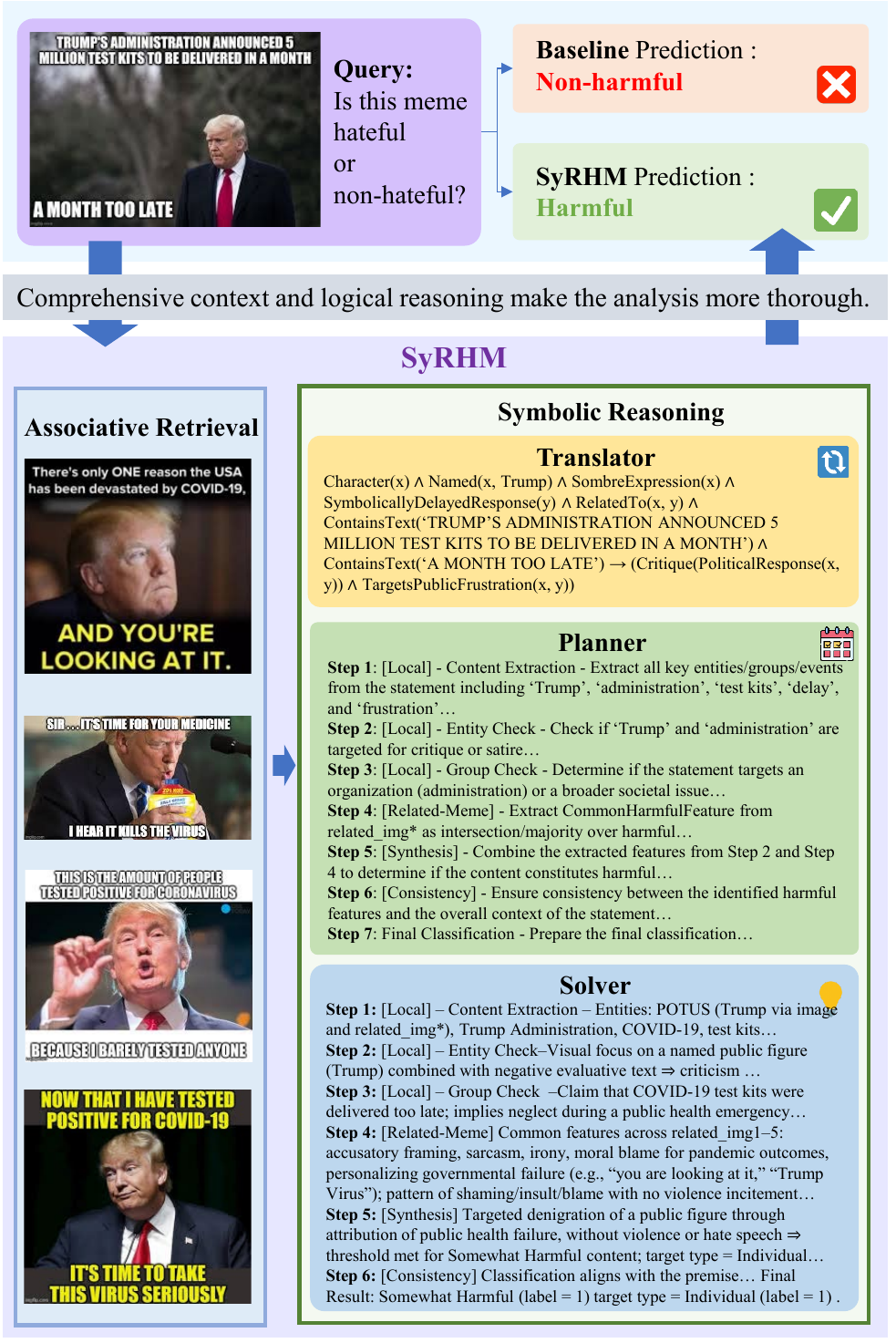}
\caption{Process Visualization of covid\_memes\_5560 from HarM. The meme criticizes delayed COVID-19 test kit delivery by Trump's administration. While the baseline model fails to detect harm, SyRHM identifies the harmful intent by leveraging retrieved contexts and analyzing sarcasm and blame-oriented framing.}
\label{visual2}
\end{figure*}

As shown in Figure~\ref{visual1} and Figure~\ref{visual2}, we present representative examples to complement the quantitative results and show how SyRHM uses associative retrieval and symbolic reasoning for interpretable meme classification. 

These cases highlight that SyRHM can effectively distinguish between visually similar but semantically different memes. By incorporating retrieved contextual evidence and structured symbolic analysis, SyRHM avoids overreliance on superficial visual cues and captures implicit harmful intentions that are difficult for conventional multimodal models to recognize.

\label{sec:appendix}

\end{document}